\documentclass[]{article}
\usepackage[letterpaper]{geometry}
\usepackage{mtsummit2021}
\usepackage{times}
\usepackage{url}
\usepackage{latexsym}
\usepackage{booktabs}
\usepackage{graphicx}
\usepackage{natbib}
\usepackage{layout}
\usepackage{xspace}
\usepackage{multirow}
\usepackage{hyperref}
\usepackage{comment}
\usepackage{caption}
\usepackage[dvipsnames]{xcolor}

\newcommand{\en}{\emph{en}}
\newcommand{\yo}{\emph{yo}}

\newcommand*{\yoruba}{Yor\`ub\'a\xspace}
\newcommand*{\menyo}{MENYO-20k\xspace}

\begin{document}

\title{\bf \menyo: A Multi-domain English--\yoruba Corpus  for Machine Translation and Domain}
\title{The Effect of Domain and Diacritics in \yoruba--English Neural Machine Translation}

\author{

\name{\bf David Ifeoluwa Adelani\thanks{* Equal contribution to the work}} \hfill  \addr{didelani@lsv.uni-saarland.de}\\
\addr{Spoken Language Systems Group (LSV), Saarland University, Germany \& Masakhane NLP}
\AND
    \name{\bf Dana Ruiter$^*$} \hfill \addr{druiter@lsv.uni-saarland.de}\\
        \addr{Spoken Language Systems Group (LSV), Saarland University, Germany}
\AND
       \name{\bf Jesujoba O. Alabi$^*$} \hfill \addr{jalabi@mpi-inf.mpg.de}\\
        \addr{Max Planck Institute for Informatics, Saarbrücken, Germany \& Masakhane NLP}
\AND
        \name{\bf Damilola Adebonojo} \hfill \addr{iyayorubagidi@gmail.com }\\
        \addr{ Alamoja Yoruba \& Masakhane NLP}
\AND
        \name{\bf Adesina Ayeni} \hfill \addr{info@yobamoodua.org}\\
        \addr{Yobamoodua Cultural Heritage (YMCH)}
\AND
\name{\bf Mofe Adeyemi} \hfill \addr{mofetoluwa@outlook.com}\\
        \addr{Defence Space Administration, Abuja, Nigeria \& Masakhane NLP}
\AND
      \name{\bf Ayodele Awokoya} \hfill \addr{ayodeleawokoya@gmail.com}\\
        \addr{University of Ibadan, Nigeria \& Masakhane NLP}
\AND
      \name{\bf Cristina Espa{\~n}a-Bonet} \hfill \addr{cristinae@dfki.de}\\
        \addr{DFKI GmbH, Saarland Informatics Campus, Saarbrücken, Germany}
}

\maketitle
\pagestyle{empty}

\begin{abstract}
\vspace{2mm}
  Massively multilingual machine translation (MT) has shown impressive capabilities, including zero and few-shot translation between low-resource language pairs. However, these models are often evaluated on high-resource languages with the assumption that they generalize to low-resource ones. The difficulty of evaluating MT models on low-resource pairs is often due to lack of standardized evaluation datasets. In this paper, we present \menyo, the first multi-domain parallel corpus with 
  a special focus on clean orthography
  for \yoruba--English with standardized train-test splits for benchmarking. We provide several neural MT benchmarks and compare them to the performance of popular pre-trained (massively multilingual) MT models both for the heterogeneous test set and its subdomains. Since these pre-trained models use huge amounts of data with uncertain quality, we also analyze the effect of diacritics, a major characteristic of \yoruba, in the training data. We investigate how and when this training condition affects the final quality and intelligibility of a translation.
  Our models outperform massively multilingual models such as Google ($+8.7$ BLEU) and Facebook M2M ($+9.1$ BLEU) when translating to \yoruba, setting a high quality benchmark for future research.

\end{abstract}

\section{Introduction}
\label{s:intro}
Neural machine translation (NMT) achieves high quality performance when large amounts of parallel sentences are available 
\citep{WMT:2020}.
Large and freely-available parallel corpora do exist for a small number of high-resource pairs and domains. However, 
for low-resource languages such as \yoruba ($yo$), one can only find few thousands of parallel sentences online\footnote{\url{http://opus.nlpl.eu}}.
In the best-case scenario, i.e. some amount of parallel data exists,
one can use the Bible ---the Bible is the most available resource for low-resource languages~\citep{Resnik1999TheBA}--- and JW300~\citep{agic-vulic-2019-jw300}. Notice that both corpora belong to the religious domain and they do not generalize well to popular domains such as news and daily conversations. 

In this paper, we address this problem for the \yoruba--English ($yo$--$en$) language pair by creating a multi-domain parallel dataset, \menyo, which we make publicly available\footnote{\url{https://github.com/uds-lsv/menyo-20k_MT}} with CC BY-NC 4.0 licence. 
It is a heterogeneous dataset that comprises texts obtained from news articles, TED talks, movie and radio transcripts, science and technology texts, and other short articles curated from the web and translated by professional translators.  
Based on the resulting train-development-test split, we provide a benchmark for the $yo$--$en$\ translation task for future research on this language pair. This allows us to properly evaluate the generalization of MT models trained on JW300 and the Bible on new domains. We further explore transfer learning approaches that can make use of a few thousand sentence pairs for domain adaptation. Finally, we analyze the effect of \yoruba diacritics on the translation quality of pre-trained MT models, discussing in details how this affects the understanding of the translated text especially in the \en--\yo  \ direction. We show the benefit of automatic diacritic restoration in addressing the problem of noisy diacritics.

\section{The \yoruba Language}
\label{s:yoruba}

The \yoruba language is the third most spoken language in Africa, and it is native to south-western Nigeria and the Republic of Benin. It is one of the national languages in Nigeria, Benin and Togo, and spoken across the West African regions. The language belongs to the Niger-Congo family, and it is spoken by over 40 million native speakers~\citep{Ethnologue2019}. 

\yoruba has 25 letters without the Latin characters c, q, v, x and z, and with additional characters {\d e}, gb, {\d s} , {\d o}. \yoruba is a tonal language with three tones: low, middle and high. These tones are represented by the grave (e.g. ``\`{a}
''), optional macron (e.g. ``\={a}'') and acute (e.g. ``\'{a}'') accents respectively.  These tones are applied on vowels and syllabic nasals, but the mid tone is usually ignored in writings. The tone information and underdots are important for the correct pronunciation of words.  Often, articles written online, including news articles such as BBC\footnote{\url{https://www.bbc.com/yoruba}}  
ignore diacritics. Ignoring diacritics makes it difficult to identify or pronounce words except when they are embedded in context. For example, \textit{{\` e}d{\` e}} (language), \textit{{e}d{\' e}} (crayfish), \textit{{\d e}d{\d e}} (a town in Nigeria), \textit{{\d{\` e}}d{\d{e}}} (trap) and \textit{{\d{\` e}}d{\d{\` e}}} (balcony) will be mapped to \textit{ede} without diacritics. 

Machine translation might be able to learn to disambiguate the meaning of words and generate correct English even with un-diacriticized \yoruba. However, one cannot generate correct \yoruba if the training data is un-diacriticized. One of the purposes of our work is to build a corpus with correct and complete diacritization in several domains.

\section{\menyo}
\label{s:collection}
The dataset collection
was motivated by the inability of machine translation models trained on JW300 to generalize to new domains~\citep{nekoto-etal-2020-participatory}. Although \citet{nekoto-etal-2020-participatory} evaluated this for \yoruba with surprisingly high BLEU scores, the evaluation was done on very few examples from the COVID-19 and TED Talks domains with 39 and 80 sentences respectively. Inspired by the FLoRes dataset for Nepali and Sinhala~\citep{guzman-etal-2019-flores}, we create a high quality test set for \yoruba-English with few thousands of sentences in different domains to check the quality of industry MT models, pre-trained MT models, and MT models based on popular corpora such as JW300 and the Bible. 

\subsection{Dataset Collection for MENYO-20k}
\autoref{tab:data_source} summarizes the texts collected, their source, the original language of the texts and the number of sentences from each source. 
We collected both parallel corpora freely available on the web (e.g JW News) and monolingual corpora we are interested in translating (e.g. the TED talks) to build the \menyo corpus. The JW News is different from the JW300 since they contain only news reports, and we manually verified that they are not in JW300. Some few sentences were donated by professional translators such as ``short texts'' in \autoref{tab:data_source}. Our curation followed two steps: (1) translation of monolingual texts crawled from the web by professional translators;
(2) verification of translation, orthography and diacritics for parallel texts obtained online and translated. Texts obtained from the web that were judged by native speakers being high quality were verified once, the others were verified twice. 
The verification of translation and diacritics was done by professional translators and volunteers who are native speakers. 

\autoref{tab:data_source} on the right (top) summarizes the figures for the \menyo dataset with 20,100 parallel sentences split into 10,070 training sentences, 3,397 development sentences, and 6,633 test sentences. The test split contains 6 domains, 3 of them have more than 1000 sentences and can be used as domain test sets by themselves.

\begin{table}[t]
\footnotesize
\begin{center}
\scalebox{0.95}{
\begin{tabular}{p{20mm}lr}
\toprule
\textbf{Data name} & \textbf{Source} & \textbf{No. Sent.}  \\
\midrule
\multicolumn{2}{l}{\textbf{source language: en-yo}} \\
JW News & \url{jw.org/yo/iroyin} & 3,508   \\
VON News & \url{von.gov.ng} & 3,048   \\
GV News &\url{globalvoices.org} & 2,932   \\
\yoruba Proverbs & \url{@yoruba\_proverbs} & 2,700   \\
Movie Transcript & ``Unsane'' on YouTube & 774 \\
UDHR & \url{ohchr.org} & 100 \\
ICT localization & from \yoruba translators & 941   \\
Short texts & from \yoruba translators & 687 \\
\multicolumn{2}{l}{\textbf{source language: en}} \\
TED talks &\url{ted.com/talks} & 2,945          \\
Out of His Mind & from the book author & 2,014   \\
Radio Broadcast & from Bond FM Radio & 258 \\
CC License & Creative Commons & 193 \\
\midrule
Total &  & 20,100  \\ 
\bottomrule
\end{tabular}
}
\quad
\scalebox{0.95}{
\begin{tabular}{p{15mm}rrr}
\toprule
 &  \multicolumn{3}{c}{\textbf{Number of Sentences}}  \\
\textbf{Domain}  & \textbf{Train. Set} & \textbf{Dev. Set} & \textbf{Test Set}  \\
\midrule
\emph{MENYO-20k} \\
\textbf{News} & 4,995 & 1,391  & 3,102    \\
\textbf{TED Talks} & 507 & 438 & 2,000   \\
\textbf{Book} &-- & 1,006 & 1,008           \\
\textbf{IT} & 356 & 312 & 273  \\
\textbf{\yoruba Proverbs} & 2,200 & 250 & 250            \\
\textbf{Others} & 2,012 & -- &  --           \\
\midrule
\multicolumn{3}{l}{\emph{Standard (religious) corpora}} \\
\textbf{Bible} & 30,760& -- & --            \\
\textbf{JW300} & 459,871 & -- & --            \\
\midrule
\emph{TOTAL} & 500,701 & 3,397 & 6,633 \\ 
\bottomrule
\end{tabular}
 }
\end{center}
\footnotesize
  \caption{\textbf{Left:} Data collection. \textbf{Right:} \menyo domains and training, development and test splits (top); figures for standard corpora used in this work (bottom).}
     \label{tab:data_source}
\end{table}

\subsection{Other Corpora for \yoruba and English}
\paragraph{Parallel corpora} For our experiments, we use two widely available parallel corpora from the religion domain, Bible and JW300 (\autoref{tab:data_source}, bottom). 
The parallel version of the Bible is not available, so we align the verses from the New International Version (NIV) for English and the Bible Society of Nigeria version (BSN) for \yoruba. 
After aligning the verses, we obtain 30,760 parallel sentences. 
Also, we download the JW300 parallel corpus which is available for a large variety of low-resource language pairs. It has parallel corpora from English to $343$ languages containing religion-related texts. From the JW300 corpus, we get $459,871$ sentence pairs already tokenized with \emph{Polyglot}\footnote{\url{https://github.com/aboSamoor/polyglot}} \citep{ramialrfou}. 

\paragraph{Monolingual Corpora}
We make use of additional monolingual data to train the semi-supervised MT model using back-translation. The \yoruba monolingual texts are from the \yoruba embedding corpus~\citep{alabi-etal-2020-massive}, one additional book (``Ojowu'') with permission from the author, JW300-\yo, and Bible-\yo. 
We only use \yoruba texts that are properly diacritized. In order to keep the topics in the \yoruba and English monolingual corpora close, we choose two Nigerian news websites (The Punch Newspaper\footnote{\url{https://punchng.com}} and Voice of Nigeria \footnote{\url{https://von.gov.ng}})
for the English monolingual corpus. The news scraped covered categories such as politics, business, sports and entertainment. Overall, we gather 475,763 monolingual sentences from the website.

\subsection{Dataset Domain Analysis}
\label{s:dataset_domain_analysis}
\menyo is, on purpose, highly heterogeneous. In this section we analyze the differences and how its (sub)domains depart from the characteristics of the commonly used \yoruba--English corpora for MT. 

Characterizing the domain of a dataset is a difficult task. Some metrics previously used need either large corpora or a characteristic vocabulary of the domain~\citep{beyer-etal-2020-embedding,EspanaBonetEtal:2020}. Here, we do not have these resources and we report the overlapping vocabulary between training and test sets and the perplexity observed in the test sets when a language model (LM) is trained on the MT training corpora.

\begin{figure*}[t]
    \centering
    \includegraphics[width=0.9\linewidth, trim={0 3.5em 0 0},clip]{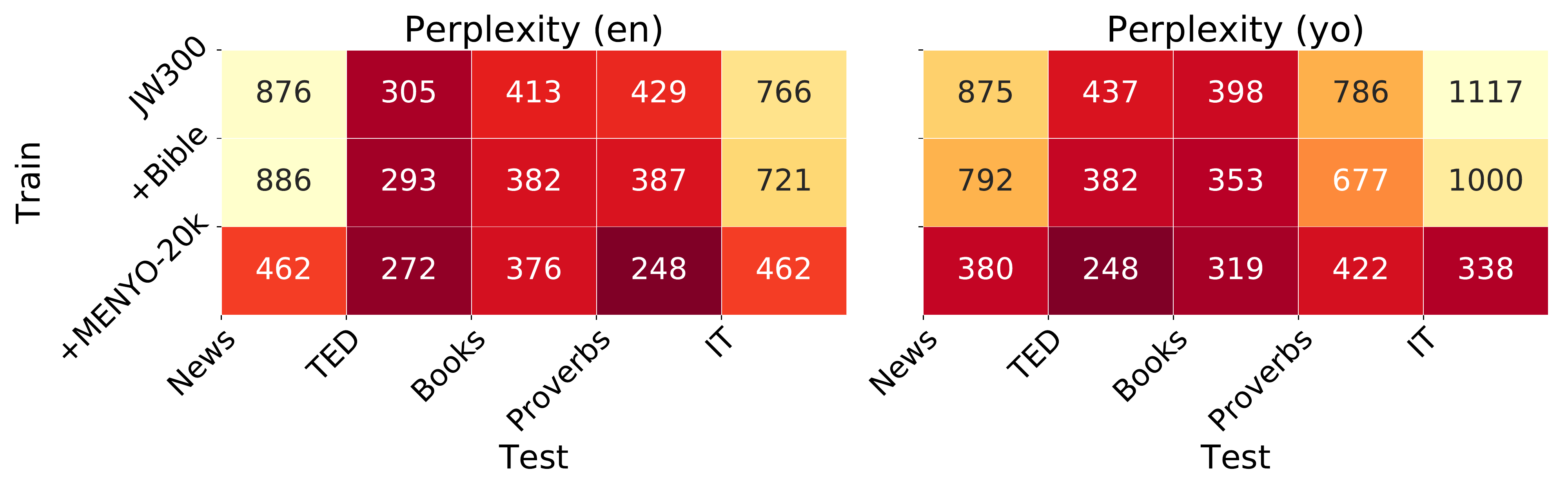}
    \smallskip
    \includegraphics[width=0.9\linewidth]{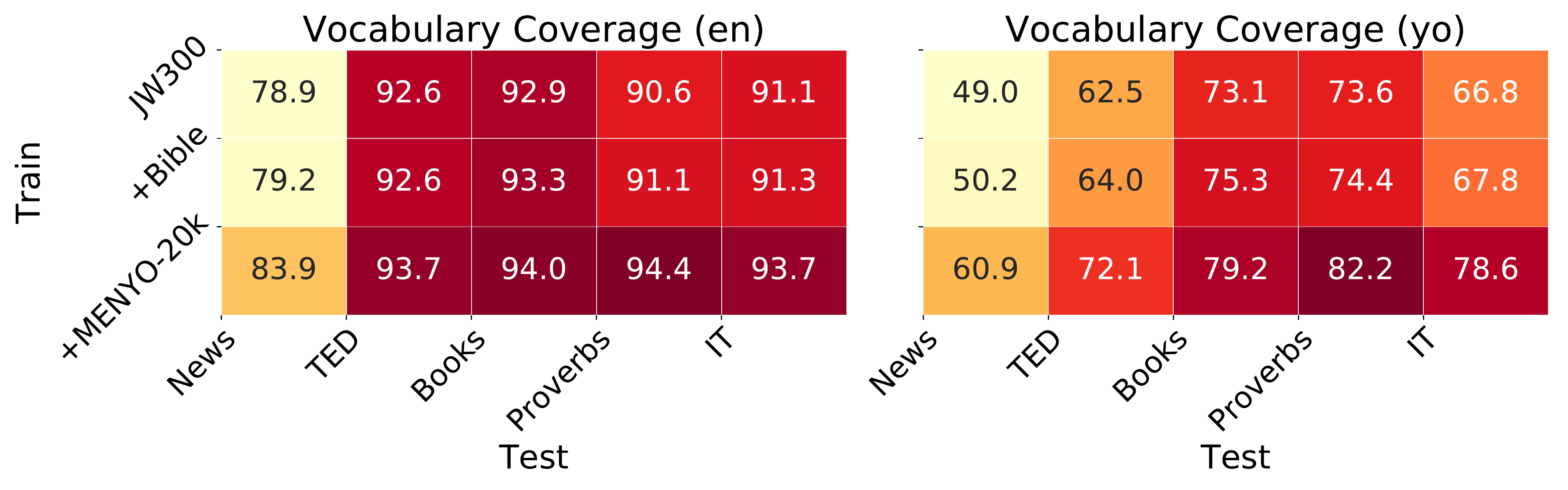}
    \caption{\textbf{Top:} Perplexities of KenLM 5-gram language model learned on different training corpora and tested on subsets of \menyo for English (left) and \yoruba (right) respectively. \textbf{Bottom:} Vocabulary coverage (\%) of different subsets of the \menyo test set per training sets for English (left) and \yoruba (right).}
    \label{f:pps}
\end{figure*}

In order to estimate the perplexities, we train a language model of order 5 with KenLM \citep{heafield-2011-kenlm} on each of the 3 training data subsets: JW300 (named C2 for short in tables), JW300+Bible (C3), JW300+Bible+\menyo (C4). 
Following NMT standard processing pipelines (see \autoref{ss:settings}), we perform byte-pair encoding (BPE) \citep{sennrich2016neural} on the corpora to avoid a large number of out-of-vocabulary tokens which, for small corpora, could alter the LM probabilities.
For each of the resulting language models, we evaluate their average \textbf{perplexity} on the different domains of the test set to evaluate \emph{compositional} domain differences (\autoref{f:pps}, top).
As expected, the average perplexity drops when adding more training data. Due to the limited domain of both JW300 and Bible, a literary style close to the Books domain, the decrease in perplexity is small when adding additional Bible data to JW300, namely $-8\%$\ (\en) and $-11\%$\ (\yo).  Interestingly, both JW300 and Bible also seem to be close to the TED domain (1st and 2nd lowest perplexities for \en\ and \yo\ respectively), which may be due to discourse/monologue content in both training corpora. Adding the domain-diverse \menyo corpus largely decreases the perplexity across all domains with a major decrease of $-66\%$ on IT (\yo) and smallest decrease of $-1\%$ on Books (\en). The perplexity scores correlate negatively with the resulting BLEU scores in \autoref{tab:bleu2_result}, with a Pearson's $r$ ($r$) of $-0.367$ (\en) and $-0.461$ (\yo), underlining that compositional domain differences between training and test subsets is the main factor of differences in translation quality.

Further, to evaluate \emph{lexical} domain differences, we calculate the \textbf{vocabulary coverage} (tokenized, not byte-pair encoded\footnote{We do not use byte-pair encoded data here, since, due to the nature of BPE, the vocabulary overlap would be close to 1 between all training and test sets.}) of the different domains of the test set by each of the training subsets (\autoref{f:pps}, bottom). The vocabulary coverage increases to a large extend when \menyo\ is added. However, while vocabulary coverage and average perplexities have a strong (negative) correlation, $r=-0.756$\ (\en) and $r=-0.689$\ (\yo), a high perplexity does not necessarily mean low vocabulary coverage. E.g., the vocabulary coverage of the IT domain by JW300 is high ($91\%$ for \en) despite leading to high perplexities ($765$ for \en). In general, vocabulary coverage of the test sets is less indicative of the resulting translation performance than perplexity, showing only a weak correlation between vocabulary coverage and BLEU, with $r=0.150$ and $r=0.281$ for \en\ and \yo\ respectively.

\section{Neural Machine Translation for \yoruba--English}
\label{s:MT}

\subsection{Systems}
\label{ss:systems}

\paragraph{Supervised NMT}
We use the transformer-base architecture proposed by \citet{NIPS2017_7181}
as implemented in Fairseq\footnote{\url{https://github.com/pytorch/fairseq}} \citep{ott2019fairseq}. We set the drop-out at 0.3
and batch size at $10,240$ tokens. 
For optimization, we use \emph{adam} \citep{adam_opt} with $\beta_1=0.9$ and $\beta_2=0.98$ and a learning rate of 0.0005. The learning rate has a warmup update of 4000, using label smoothed cross-entropy loss function with label-smoothing value of 0.1. 

\paragraph{Semi-supervision via iterative back-translation}
We use the best performing supervised system to translate the monolingual corpora described in \autoref{s:collection} yielding to 476$k$ back-translations. This data is used together with the original corpus to train a new system. The process is repeated until convergence.

\paragraph{Fine-tuning mT5}
We examine a transfer learning approach by fine-tuning a massively multilingual model mT5~\citep{Xue2020mT5AM}.
mT5 had been pre-trained on 6.3T tokens originating from Common Crawl in 101 languages (including \yoruba). The approach has already shown competitive results on other languages~\citep{Tang2020MultilingualTW}. In our experiments, we use mT5-base, a model with 580M parameters. We transferred all the parameters of the model including the sub-word vocabulary.

\paragraph{Publicly Available NMT Models}
We further evaluate the performance of three multilingual NMT systems: OPUS-MT \citep{TiedemannThottingal:EAMT2020}, Google Multilingual NMT (GMNMT) \citep{BapnaEtAl:2019} and Facebook's M2M-100 \citep{fan2020englishcentric} with 1.2B parameters.
All the three pre-trained models are trained on over 100 languages. While GMNMT and M2M-100 are a single multilingual model, OPUS-MT models are for each translation direction, e.g \yo--\en. 
We generate the translations of the test set using the \emph{Google Translate} interface,\footnote{\url{https://translate.google.com/}} 
and OPUS-MT 
using \emph{Easy-NMT}.\footnote{\url{https://github.com/UKPLab/EasyNMT}} For M2M-100, we make use of \emph{Fairseq}  to translate the test set.

\subsection{Experimental Settings}
\label{ss:settings}

\paragraph{Data and Preprocessing}
For the MT experiments, we use the training part of our \menyo\ corpus  and two other parallel corpora, Bible and JW300 (\autoref{s:collection}). 
For tuning the hyper-parameters, we use the development split of the multi-domain data which has $3,397$ sentence pairs and for testing the test split with $6,633$ parallel sentences.
To ensure that all the parallel corpora are in the same format, we convert the \yoruba texts in the JW300 dataset to Unicode Normalization Form Composition (NFC), the format of the \yoruba texts in the Bible and multi-domain dataset. 
Our preprocessing pipeline includes punctuation normalization, tokenization, and truecasing. For punctuation normalization and truecasing, we use the \emph{Moses} toolkit \citep{koehn-etal-2007-moses} while for tokenization, we use \emph{Polyglot}, since it is the tokenizer used in JW300. We apply joint BPE, with a vocabulary threshold of 20 and 40k merge operations.

\paragraph{Evaluation Metrics} To evaluate the models,
we use tokenized BLEU~\citep{papineni-etal-2002-bleu} score implemented in \emph{multi-bleu.perl} and confidence intervals ($p=95\%$) in the scoring package\footnote{\url{https://github.com/lvapeab/confidence_intervals}}. 
Since diacritics are applied on individual characters, we also use chrF, a character $n$-gram F1-score~\citep{popovic-2015-chrf}, for \en--\yo \ translations.

\paragraph{Automatic Diacritization}
In order to automatically diacritize Google MNMT and M2M-100 outputs for comparison, we train an automatic diacritization system using the supervised NMT setup. We use the \yoruba\  side of \menyo and JW300, which use consistent diacritization. We split the resulting corpus into train (458$k$ sentences), test (517 sentences) and development (500 sentences) portions. We apply a small BPE of 2k merge operations to the data. We apply noise on the diacritics by $i)$ randomly removing a diacritic with probability $p=0.3$ and $ii)$ randomly replacing a diacritic with $p=0.3$. The corrupted version of the corpus is used as the source data, and the NMT model is trained to reconstruct the original diacritics. On the test set, where the corrupted source has a BLEU (precision) of $19.0$ ($29.8$), reconstructing the diacritics using our system lead to a BLEU (precision) of $87.0$ ($97.1$), thus a major increase of $+68.0$ ($+67.3$) respectively.

\subsection{Automatic Evaluation}
\label{ss:results}

\paragraph{Internal Comparison} We train four basic NMT engines on different subsets of the training data: Bible (C1), JW300 (C2), JW300+Bible (C3) and JW300+Bible+\menyo (C4). Further, we analyse the effect of fine-tuning for in-domain translation. For this, we fine-tune the converged model trained on JW300+Bible on \menyo (C3+Transfer) and, similarly, we fine-tune the converged model trained on JW300+Bible+\menyo on \menyo (C4+Transfer). This yields six NMT models in total for \en--\yo \ and \yo--\en\ each.
Their translation performance is evaluated on the complete \menyo\ test set (\autoref{tab:bleu1}, top) and later we analyze in-domain translation in \autoref{tab:bleu2_result}.

As expected, the BLEU scores obtained after training on Bible only (C1) are low, with BLEU $2.2$ and $1.4$ for \en--\yo \ and \yo--\en\ respectively, which is due to its small amount of training data. Training on the larger JW300 corpus (C2) leads to higher scores of BLEU $7.5$ (\en--\yo ) and $9.6$ (\yo--\en), while combining it with Bible (C3) only leads to a small increase of BLEU $+0.6$  and $+1.2$ for \en--\yo \ and \yo--\en\ respectively. 
When further adding \menyo\ (C4) to the training data, the translation quality increases by $+2.8$ (\en--\yo ) and $+3.2$ (\yo--\en). When, instead of adding \menyo\ to the training pool, it is used to fine-tune the converged JW300+Bible model, (C3+Transfer) the increase in BLEU over JW300+Bible is even larger for \en--\yo \ (BLEU $+4.2$), which results in an overall top-scoring model with BLEU $12.3$. For \yo--\en\ fine-tuning  is slightly less effective (BLEU $13.2$) than simply adding \menyo\ to the training data (BLEU $14.0$). As seen in \autoref{s:dataset_domain_analysis}, perplexities and vocabulary coverage in English are not as distant among training/test sets as in \yoruba, so the fine-tuning step resulted less efficient. 

When we use the \menyo dataset to fine-tune the converged JW300+Bible+ \menyo model (C4+Transfer) we observe an increase in BLEU over JW300+Bible for both translation directions: $+4.3$ for \en--\yo \ and $+3.8$ for \yo--\en. This is the best performing system and the one we use for back-translation.
\autoref{tab:bleu1} also shows the performance of the semi-supervised system (C4+Transfer+BT). After two iterations of BT, we obtain an improvement of $+3.6$ BLEU points on \yo--\en.
There is, however, no improvement in the \en--\yo \ direction probably because a significant portion of our monolingual data is based on JW300.
Finally, fine-tuning mT5 with \menyo does not improve over fine-tuning only the JW300+Bible system on \en--\yo , but it does for \yo--\en. Again, multilingual systems are stronger when used for English, and we need the contribution of back-translation to outperform the generic mT5.

\begin{table}[t]
\footnotesize
 \begin{center}
  \begin{tabular}{p{47mm}cccccc}
   \toprule
   \textbf{Model} &   \multicolumn{2}{c}{\textbf{\en--\yo }}  & \multicolumn{2}{c}{\textbf{\en--\yo $^p$}} & \textbf{\yo--\en} & \textbf{\yo--\en$^u$} \\
                  & chrF & BLEU & chrF & BLEU & BLEU & BLEU \\
   \midrule
   \emph{Internal Comparison} \\
   \textbf{C1: Bible}                 & 16.9 &  2.2$\pm$0.1 & --&-- & 1.4$\pm$0.1 & 1.6$\pm$0.1   \\
   \textbf{C2: JW300 }                & 29.1 &  7.5$\pm$0.2 & --&-- & 9.6$\pm$0.3 & 9.3$\pm$0.3    \\
   \textbf{C3: JW300{+}Bible}         & 29.8 &  8.1$\pm$0.2 & --&-- & 10.8$\pm$0.3 & 10.5$\pm$0.3    \\
   \textbf{~~~~~~~~{+}Transfer}       & 33.8 & 12.3$\pm$0.3 & --&-- &13.2$\pm$0.3 & 13.9$\pm$0.3   \\
   \textbf{C4: JW300{+}Bible{+}\menyo}& 32.5 & 10.9$\pm$0.3 & --&-- & 14.0$\pm$0.3 & 14.0$\pm$0.3     \\
   \textbf{~~~~~~~~{+}Transfer}       & 34.3 & \underline{\textbf{12.4}}$\pm$0.3 &-- &--   & 14.6$\pm$0.3 &   --  \\
   \textbf{~~~~~~~~~~{+} BT}          & 34.6 & 12.0$\pm$0.3 &-- &--  & \underline{18.2$\pm$0.4} & -- \\
   \textbf{mT5: mT5-base{+}Transfer}  & 32.9 & 11.5$\pm$0.3 & --&-- & 16.3$\pm$0.4 & 16.3$\pm$0.4 \\
      \midrule
    \emph{External Comparison} \\
    \textbf{OPUS-MT}           & -- & -- &-- & --                    &  \multicolumn{1}{c}{5.9$\pm$0.2} & --  \\
    \textbf{Google GMNMT}      & 18.5 & 3.7$\pm$0.2 & 34.4 & 10.6$\pm$0.3  &  \multicolumn{1}{c}{\textbf{22.4}$\pm$0.5} & --        \\
    \textbf{Facebook M2M-100}  & 15.8 & 3.3$\pm$0.2 & 25.7 & 6.8$\pm$0.3   & \multicolumn{1}{c}{4.6$\pm$0.3}     & -- \\
   \bottomrule
  \end{tabular}
  \footnotesize
  \caption{Tokenized BLEU with confidence intervals ($p=95\%$) and chrF scores over the full test for NMT models trained on different subsets of the training data C$_i$ (top) and performance of external systems (bottom). For \yoruba, we analyse the effect of diacritization: \en--\yo $^p$ applies an in-house diacritizer on the translations obtained from \textbf{p}re-trained models and \yo--\en$^u$ reports results using \textbf{u}ndiacritized \yoruba texts as source sentences for training (see text).
  Top-scoring results per block are underlined and globally boldfaced.
  }
  \label{tab:bleu1}
 \end{center}
\end{table}

\paragraph{External Comparison}
We evaluate the performance of the open source multilingual engines introduced in the previous section
on the full test set (\autoref{tab:bleu1}, bottom). 
\textbf{OPUS-MT}, while having no model available for \en--\yo , achieves a BLEU of $5.9$ for \yo--\en. Thus, despite being trained on JW300 and other available {\yo--\en} corpora on OPUS, it is largely outperformed by our NMT model trained on JW300 only (BLEU $+3.7$). This may be caused by some of the noisy corpora included in OPUS (like CCaligned)
, which can depreciate the translation quality.

Facebook's \textbf{M2M-100}, is also largely outperformed even by our simple JW300 baseline by 5 BLEU points in both translation directions.
A manual examination of the \en--\yo\ LASER extractions used to train M2M-100 shows that these are very noisy similar to the findings of \citet{qualityAtAGlance}, which explains the poor translation performance. 

Google, on the other hand, obtains impressive results with \textbf{GMNMT} for the \yo--\en\ direction, with BLEU $22.4$. The opposite direction \en--\yo , however, shows a significantly lower performance (BLEU $3.7$), being outperformed even by our simple JW300 baseline (BLEU $+3.8$). The difference in performance for English can be attributed to the highly multilingual but English-centric nature of the Google MNMT model. As already noticed by \citet{BapnaEtAl:2019}, low-resourced language pairs benefit from multilinguality when translated into English, but improvements are minor when translating into the non-English language. For the other translation direction, \en--\yo , we notice that lots of diacritics are lost in Google translations, damaging the BLEU scores. 
Whether this drop in BLEU scores really affects understanding or not is analyzed via a human evaluation (Section \ref{ss:manual}). 

\paragraph{Diacritization}
Diacritics are important for \yoruba embeddings \citep{alabi-etal-2020-massive}. However, they are often ignored in popular multilingual models (e.g. multilingual BERT \citep{devlin-etal-2019-bert}), and not consistently available in training  corpora and even test sets.
In order to investigate whether the diacritics in \yoruba MT can help to disambiguate translation choices, we additionally train  \yo--\en$^u$ equivalent models on \textbf{u}ndiacritized JW300, JW300+Bible and JW300+Bible+\menyo (\autoref{tab:bleu1}, indicated as \yo--\en$^u$\ in comparison to the ones with diacritics \yo--\en). Since one cannot generate correct \yoruba text when training without diacritics, \en--\yo$^u$ \ systems are not trained. Alternatively, we restore diacritics using our in-house diacritizer in the output of open source models that produce undiacritized text.

Results for \yo--\en\ are not conclusive. Diacritization is useful when only out-of-domain data is used in training (JW300, JW300+Bible\footnote{We do not consider Bible alone. Due to its small data size, the BLEU scores are less indicative.} for testing on \menyo).
In this case, the domain of the training data is very different from the domain of the test set, and disambiguation is needed not to bias all the lexicon towards the religious domain. When we include in-domain data (JW300+Bible+\menyo), both models perform equally well, with BLEU $14.0$ for both diacritized and undiacritized versions. Diacritization is not needed when we fine-tune the model with data that shares the domain with the test set (JW300+Bible+Transfer), BLEU is $13.2$ for the diacritized version vs. BLEU $13.9$ for the undiacritized one.

In practice, this means that, when training data is far from the desired domain, investing work for a clean diacritized \yoruba\ source input can help improve the translation performance. When more data is present, the diacritization becomes less important, since context is enough for disambiguation.

When \yoruba is the target language, diacritization is always needed. An example is the low automatic scores GMNMT (BLEU 3.7, chrF $18.5$) and M2M-100 (BLEU 3.3, chrF $15.8$) reach for \en--\yo \  translation. \autoref{tab:bleu1}-bottom (indicated as \en--\yo$^p$) show the improvements after automatically restoring the diacritics, namely $BLEU +6.9$ points, chrF $+15.9$ for GMNMT; and $+3.5$ and $+9.9$ for M2M-100. Even if the diacritizer is not perfect, diacritics do not seem enough to get state-of-the-art results according to automatic metrics: fine-tuning with high quality data (C4+Transfer+BT, chrF $34.6$) is still better than using huge but unadapted systems.

\begin{table}[t!]
\footnotesize
\centering
  \begin{tabular}{l rrrrr  rrrrr}
    \toprule
      & \multicolumn{5}{c}{\textbf{\en--\yo }}  & \multicolumn{5}{c}{\textbf{\yo--\en}}
    \\
        \cmidrule(l){2-6} \cmidrule(l){7-11}

     & \textbf{Prov.} & \textbf{News} & \textbf{TED} & \textbf{Book} & \textbf{IT} & \textbf{~~~~Prov.} & \textbf{News} & \textbf{TED} & \textbf{Book} & \textbf{IT}
    \\
    \cmidrule(l){1-11}
    \textbf{C1}           & 0.8 &  1.7 & 3.1 &  3.4 & 1.5 & 1.1 & 0.9 &  2.1 &  2.4 & 0.9   \\
    \textbf{C2}           & 2.2 & 6.4 & 9.8 &  9.8 & 4.8 & 2.6 & 8.4 & 13.1 &  9.6 & 7.0   \\
    \textbf{C3}   & 3.5 & 6.7 & 10.7& 11.3 & 4.9 & 4.8 & 9.5 & 14.4 & 10.9 & 7.8   \\
    \textbf{~~~~~{+}Transfer}   & 9.0 & 10.2 & ~\textbf{16.1} & \textbf{15.0} & 11.8 & 8.6 & 12.5 & 16.8 & 10.8 & 9.7   \\ 
    \textbf{C4}     & 7.0 & 10.0 & 12.3 & 11.5 & 10.5 & 8.7 & 13.5 & 16.7 & 11.6 & 12.4   \\
    \textbf{~~~~~{+}Transfer}     & \textbf{10.3} & 10.9 & 15.1 & 13.2 & \textbf{13.6} & \textbf{9.3} & 14.0 & 17.8 & 11.9 & 13.7   \\
    \textbf{~~~~~~~~~{+}BT} & 7.5 & 11.4 & 12.9 & 14.5 & 9.7 & 7.9 & \textbf{18.6} & \textbf{20.6} & \textbf{13.3} & \textbf{16.4} \\
    \textbf{mT5{+}Transfer} & 3.8 & \textbf{11.2} & 13.1 & 11.8 & 7.9 & 6.0 & 16.4 & 18.9 & 13.1 & 15.1 \\
    \bottomrule
  \end{tabular}
  \vspace{-1mm}
  \caption{Tokenized BLEU over different domains of the test set for NMT models trained on different subsets of the training data, with top-scoring results per domain in bold.}
        \label{tab:bleu2_result}
\end{table}

\paragraph{Domain Differences}
In order to analyze the domain-specific performance of the different NMT models, we evaluate each model on the different domain subsets of the test set (\autoref{tab:bleu2_result}). 
The Proverb subset is especially difficult in both directions, as it shows the lowest translation performance across all domains, i.e. maximum BLEU of $9.04$ (\en--\yo ) and $8.74$ (\yo--\en). This is due to the fact that proverbs often do not have literal counterparts in the target language, thus making them especially difficult to translate. The TED domain is the best performing test domain, with maximum BLEU of $16.1$ (\en--\yo ) and $16.8$ (\yo--\en). This can be attributed to the decent base coverage of the TED domain by JW300 and Bible together (monologues) with the additional TED domain data included in the \menyo\ training split ($507$ sentence pairs).
Also, most BLEU results are on line with the LM perplexity results and conclusions drawn in \autoref{s:dataset_domain_analysis}. Due to the closeness of Bible and JW300 to the book domain, we see only small improvements of BLEU on this domain, i.e. $+0.2$ (\en--\yo ) and $+0.7$ (\yo--\en), when adding \menyo\ (C4) to the JW300+Bible (C3) training data pool. On the other hand, the IT domain benefits the most from the additional \menyo\ data, with major gains of BLEU $+5.5$ (\en--\yo ) and $4.6$ (\yo--\en), owing to the introduction of IT domain content in the \menyo\ training data ($\sim1k$ sentence pairs), which is completely lacking in JW300 and Bible.

\begin{table}[t!]
 \footnotesize
 \centering
  \begin{tabular}{p{17mm}rrrp{3mm}rrr}
    \toprule
     & \multicolumn{3}{c}{\textbf{\en--\yo }}  & & \multicolumn{3}{c}{\textbf{\yo--\en}} 
    \\
    \textbf{Task} & \textbf{C4+Trf} & \textbf{C4+Trf+BT} & \textbf{GMNMT} & & \textbf{mT5+Trf} & \textbf{C4+Trf+BT} & \textbf{GMNMT} 
    \\
    \midrule
    Adequacy         & 3.12* & 3.58 & \textbf{3.69} & & 3.42* & 3.41* & \textbf{4.02}    \\
    Fluency          & \textbf{4.57*} & 4.49* & 3.74 & & 4.39* & 4.18* & \textbf{4.71} \\
    Diacritics acc.  & \textbf{4.91*} & 4.90* & 1.74 & & \_ & \_ & \_ \\
    \bottomrule
  \end{tabular}
  \caption{Human evaluation for \en--\yo  \ and \yo--\en \ MT models (C4+Transfer (C4+Trf), C4+Trf+BT, mT5+Trf, and GMNMT) in terms of Adequacy, Fluency and Diacritics prediction accuracy. The rating that is significantly different from GMNMT is indicated by * (T-test $p<0.05$)}. 
  \label{tab:human}
\end{table}

\subsection{Human Evaluation}
\label{ss:manual}
To have a better understanding of the quality of the translation models and the intelligibility of the translations, we compare three top performing models in \en--\yo \ and \yo--\en. For \en--\yo , we use \textbf{C4+Transfer}, \textbf{C4+Transfer+BT} and \textbf{GMNMT}. Although GMNMT is not the third best system according to BLEU  (\autoref{tab:bleu1}), we are interested in the study of diacritics in translation quality and intelligibility.
For the \yo--\en, we choose  \textbf{C4+Transfer+BT}, \textbf{mT5+Transfer} and \textbf{GMNMT} being the 3 models with the highest BLEU scores on \autoref{tab:bleu1}. 

We ask 7 native speakers of \yoruba that are fluent in English to rate the adequacy, fluency and diacritic accuracy in a subset of test sentences. Four of them rated the \en--\yo  \ translation direction and the others rated the opposite direction \yo--\en. We randomly select $100$ sentences within the outputs of the six systems and duplicate $5$ of them to check the intra-agreement consistency of our raters. Each annotator is then asked to rate 105 sentences per system on a $1-5$ Likert scale for each of the features (for English, diacritic accuracy cannot be evaluated). We calculate the agreement among raters using Krippendorff's $\alpha$. 
The inter-agreement per task is $0.44$ (adequacy), $0.40$ (fluency) and $0.87$ (diacritics) for \yoruba, and $0.71$ (adequacy), $0.55$ (fluency) for English language. We observe that a lot of raters often rate the fluency score for many sentences with the same values (e.g 4 or 5), which results to a lower Krippendorff's $\alpha$ for fluency. 
The intra-agreement for the four \yoruba raters are $0.75$, $0.91$, $0.66$, and $0.87$, while the intra-agreement for the three English raters across all evaluation tasks are $0.92$, $0.71$, and $0.81$.  

For \yo--\en, our evaluators rated on average GMNMT to be the best in terms of adequacy ($4.02$ out of 5) and fluency ($4.71$), followed by mT5+Transfer, which shows that fine-tuning massively multilingual models also benefits low resource languages MT especially in terms of fluency ($4.39$). 
This contradicts the results of the automatic evaluation which prefers C4+Transfer+BT over mT5+Transfer.

For \en--\yo , GMNMT is still the best in terms of adequacy ($3.69$) followed by C4+Transfer+BT, but performs the worst in terms of fluency and diacritics prediction accuracy. So, the bad quality of the diacritics affects fluency and drastically penalises automatic metrics such as BLEU, but does not interfere with the intelligibility of the translations as shown by the good average adequacy rating. Automatic diacritic restoration for \yoruba~\citep{orife2018adr,Orife2020ImprovingYD} can therefore be very useful to improve translation quality. 
C4+Transfer and C4+Transfer+BT perform similarly with high scores in terms of fluency and near perfect score in diacritics prediction accuracy ($4.91 \pm 0.1$) as a result of being trained on cleaned corpora. 

\section{Related Work}
\label{s:relatedwork}

In order to make MT available for a broader range of linguistic communities, recent years have seen an effort in creating new \textbf{parallel corpora} for low-resource language pairs. 
Recently, \citet{guzman-etal-2019-flores} provided novel supervised, semi-supervised and unsupervised benchmarks for Indo-Aryan languages \{Sinhala,Nepali\}--English on an evaluation set of professionally translated sentences sourced from the Sinhala, Nepali and English Wikipedias. 

Novel parallel corpora focusing on \textbf{African languages} cover South African languages (\{Afrikaans, isiZulu, Northern Sotho, Setswana, Xitsonga\}--English) \citep{groenewald-fourie-2009-introducing} with MT benchmarks evaluated in \citet{martinus2019focus}, as well as multidomain (News, Wikipedia, Twitter, Conversational) Amharic--English \citep{hadgu2020evaluating} and multidomain (Government, Wikipedia, News etc.) Igbo--English \citep{ezeani2020igboenglish}. 
Further, the LORELEI project \citep{strassel-tracey-2016-lorelei} has created parallel corpora for a variety of low-resource language pairs, including a number of Niger-Congo languages such as \{isiZulu, Twi, Wolof, \yoruba\}--English. However, these are not open-access.
On the contrary, Masakhane \citep{nekoto-etal-2020-participatory} is an ongoing participatory project focusing on creating new freely-available parallel corpora and MT benchmark models for a large variety of African languages. 

While creating parallel resources for low-resource language pairs is one approach to increase the number of linguistic communities covered by MT, this does not scale to the sheer amount of possible language combinations. Another research line focuses on \textbf{low-resource MT} from the modeling side, developing methods which allow a MT system to learn the translation task with smaller amounts of supervisory signals. This is done by exploiting the weaker supervisory signals in larger amounts of available monolingual data, e.g. by identifying additional parallel data in monolingual corpora \citep{artetxe2018margin,schwenk2019wikimatrix,schwenk2020ccmatrix}, comparable corpora \citep{ruiter-etal-2019-self, ruiterEtAl:2021}, or by including auto-encoding \citep{currey-etal-2017-copied} or language modeling tasks \citep{gulcehre2015using, ramachandran-etal-2017-unsupervised} during training. 
Low-resource language pairs can benefit from high-resource languages through transfer learning \citep{zoph-etal-2016-transfer}, e.g. in a zero-shot setting \citep{johnson2016google}, by using pre-trained language models \citep{lample2019cross}, or finding an optimal path of pivoting through related languages \citep{leng2019unsupervised}.
By adapting the model hyperparameters to the low-resource scenario, \citet{sennrich2019revisiting} were able to achieve impressive improvements over a standard NMT system.

\section{Conclusion}
\label{s:conclusions}

We present \menyo, a novel \en--\yo\ multi-domain parallel corpus for machine translation and domain adaptation. By defining a standardized train-development-test split of this corpus, we provide several NMT benchmarks for future research on the \en--\yo\ MT task. Further, we analyze the domain differences on the \menyo\ corpus and the translation performance of NMT models trained on religion corpora, such as JW300 and Bible, across the different domains. We show that, despite consisting of only $10k$ parallel sentences, adding the \menyo\ corpus train split to JW300 and Bible largely improves the translation performance over all domains. Further, we train a variety of supervised, semi-supervised and fine-tuned MT benchmarks on available \en--\yo\ corpora, creating a high quality baseline that outperforms current massively multilingual models, e.g. Google MNMT by BLEU $+18.8$ (\en--\yo ). This shows the positive impact of using smaller amounts of high-quality data (e.g. C4+Transfer, BLEU $12.4$) that takes into account language-specific characteristics, i.e. diacritics, over massive amounts of noisy data (Facebook M2M-100, BLEU $3.3$).  Apart from having low BLEU scores, our human evaluation reveals that models trained on low-quality diacritics (Google MNMT) suffer especially in fluency, while still being intelligible to the reader. While correctly diacritized data is vital for translating \en--\yo , it only has an impact on the quality of \yo--\en\ translation quality when there is a domain mismatch between training and testing data.

\section*{Acknowledgements}
We would like to thank Adebayo O. Adeojo, Babunde O. Popoola, Olumide Awokoya, Modupe Olaniyi, Princess Folasade, Akinade Idris, Tolulope Adelani, Oluyemisi Olaose, and Benjamin Ajibade for their support in translating English sentences to \yoruba, verification of \yoruba diacritics, and human evaluation. We thank Bayo Adebowale and `Dele `Adelani for donating their books (``Out of His Mind'', and ``Ojowu''). We thank Iroro Orife for providing the Bible corpus and \yoruba Proverbs corpus. We thank Marine Carpuat,  Mathias Müller, and the entire Masakhane NLP community for their feedback. We are also thankful to Damyana Gateva for evaluations with open-source models. This project was funded by the AI4D language dataset fellowship~\citep{Siminyu2021AI4DA}\footnote{\url{https://www.k4all.org/project/language-dataset-fellowship/}}. DIA acknowledges the support of the EU-funded H2020 project COMPRISE under grant agreement No. 3081705. CEB is partially funded by the German Federal Ministry of Education and Research under the funding code 01IW20010. The authors are responsible for the content of this publication. 

\small

\bibliographystyle{apalike}
\bibliography{mtsummit2021,anthology}

\newpage
\appendix

\section{Appendix}
\label{sec:appendix}

\begin{table*}[t]
  \footnotesize
 \begin{center}
\begin{tabular}{p{40mm}lp{15mm}r}
\toprule
\textbf{Data name} & \textbf{Source} & \textbf{Language of source} & \textbf{No. Sentences}  \\
\midrule
Jehovah Witness News & \url{jw.org/yo/iroyin}& en-yo & 3,508   \\
Voice of Nigeria News & \url{von.gov.ng} & en-yo & 3,048   \\
TED talks &\url{ted.com/talks} & en & 2,945          \\
Global Voices News &\url{yo.globalvoices.org} & en-yo  & 2,932   \\
Proverbs & \url{twitter.com/yoruba_proverbs}& yo-en & 2,700   \\
Out of His Mind Book & Obtained from  the author & en & 2,014           \\
Software localization & Obtained from Professional Translators  & en & 941          \\
Movie Transcript (``Unsane'') & \url{youtu.be/hdWP0X5msZQ} & yo-en & 774 \\
Short texts & Obtained from Professional Translators & en & 687 \\
Radio Broadcast & Transcript from Bond FM 92.9 Radio & en & 258 \\
Creative Commons License & Obtained from Professional Translators & en & 193 \\
UDHR Translation & \url{ohchr.org} & en-yo & 100 \\
\midrule
Total &  & & 20,100  \\ 
\bottomrule
\end{tabular}
\footnotesize
  \caption{Dataset collection sources with source language(s) and the number of sentences contained.}
   \label{tab:data_source_2}
\end{center}
\end{table*}

\subsection{Dataset Collection for \menyo}
\autoref{tab:data_source_2} summarizes the texts collected, their source, the original language of the texts and the number of sentences from each source. 
We collected both parallel corpora freely available on the web and monolingual corpora we are interested in translating (e.g. the TED talks) to build the \menyo corpus. Some few sentences were donated by professional translators such as ``short texts'' in \autoref{tab:data_source_2}. We provide more specific description of the data sources below.

\paragraph{Jehovah Witness News} We collected only parallel \textit{``newsroom''} (or \textit{``Ìròyìn''} in \yoruba) articles from \textit{\url{JW.org}} website to gather texts that are not in the religious domain. As shown in \autoref{tab:data_source_2}, we collected 3,508 sentences from their website, and we manually confirmed that the sentences are not in JW300. The content of the news mostly reports persecutions of Jehovah witness members around the world, and may sometimes contain Bible verses to encourage believers. 

\paragraph{Voice of Nigerian News} We extracted parallel texts from the VON website, a Nigerian Government news website that supports seven languages with wide audience in the country (Arabic, English, Fulfulde, French, Hausa, Igbo, and \yoruba). Despite the large availability of texts, the quality of \yoruba texts is very poor, one can see several issues with orthography and diacritics. We asked translators and other native speakers to verify and correct each sentence. 

\paragraph{Global Voices News} We obtained parallel sentences from the Global Voices website\footnote{\url{https://globalvoices.org}} contributed by journalists, writers and volunteers. The website supports over 50 languages, with contents mostly translated from English, French, Portuguese or Spanish. 

\paragraph{TED Talks Transcripts} We selected 28 English TED talks transcripts mostly covering issues around Africa like health, gender equality, corruption, wildlife, and social media e.g ``How young Africans found a voice on Twitter'' (see the \autoref{tab:ted_talks_topic} for the selected TED talk titles). The articles were translated by a professional translator and verified by another one. 

\paragraph{Proverbs} \yoruba has many proverbs and culturally referred to words of wisdom that are often referenced by elderly people. We obtained 2,700 sentences of parallel $yo$--$en$\ texts from Twitter.\footnote{Also available in \url{https://github.com/Niger-Volta-LTI/yoruba-text}}

\paragraph{Book} With permission from the author (Bayo Adebowale) of the ``Out of His Mind'' book, originally published in English, we translated the entire book to \yoruba and verified the diacritics.  

\paragraph{Software Localization Texts (Digital)} We obtained translations of some software documentations such as Kolibri\footnote{\url{https://learningequality.org/kolibri}} from past projects of professional translators. These texts include highly technical terms.

\paragraph{Movie Transcripts} We obtained the translation of a Nigerian movie ``Unsane'' on YouTube from the past project of a professional translator. The language of the movie is \yoruba and English, with transcription also provided in English. 

\paragraph{Other Short Texts} Other short texts like UDHR, Creative Commons License, radio transcripts, and texts were obtained from professional translators and online sources. \autoref{tab:data_source} summarizes the number of sentences obtained from each source. 

\begin{table*}[t]
\footnotesize
 \begin{center}
\begin{tabular}{|l|l|r|}
\toprule
 &  \textbf{Title} & \textbf{Topic}   \\
\midrule
1 & Reducing corruption takes a specific kind of investment  & Politics    \\
2 & How young Africans found a voice on Twitter & Technology   \\
3 & Mothers helping mothers fight HIV & Health           \\
4 & How women are revolutionizing Rwanda & Gender-equality  \\
5 & How community-led conservation can save wildlife & Wildlife            \\
6 & How cancer cells communicate - and how we can slow them down & Health \\ 
7 & You may be accidentally investing in cigarette companies & Health \\ 
8 & How deepfakes undermine truth and threaten democracy & Politics \\ 
9 & What tech companies know about your kids & Technology \\ 
10 & Facebook's role in Brexit - and the threat to democracy & Politics \\ 
11 & How we can make energy more affordable for low-income families & Energy \\ 
12 & Can we stop climate change by removing CO2 from the air? & Climate \\ 
13 & A comprehensive, neighborhood-based response to COVID-19 & Health \\ 
14 & Why civilians suffer more once a war is over & Human Rights \\ 
15 & Lessons from the 1918 flu
 & Health \\ 
16 & Refugees have the right to be protected & Human Rights \\ 
17 & The beautiful future of solar power & Energy \\ 
18 & How bees can keep the peace between elephants and humans & Wildlife \\ 
19 & Will automation take away all our jobs?
 & Technology \\ 
20 & A celebration of natural hair & Beauty \\ 
21 & Your fingerprints reveal more than you think & Technology \\ 
22 & Our immigration conversation is broken - here's how to have a better one & Politics \\ 
23 & What I learned about freedom after escaping North Korea & Politics \\ 
24 & Medical tech designed to meet Africa's needs & Health \\ 
25 & What's missing from the American immigrant narrative & Education \\ 
26 & A hospital tour in Nigeria & Health \\ 
27 & How fake news does real harm & Politics \\ 
28 & How we can stop Africa's scientific brain drain & Education \\ 
\bottomrule
\end{tabular}
\footnotesize
  \caption{TED talks titles.}
  \label{tab:ted_talks_topic}
\end{center}
\end{table*}

\end{document}